# Ring-based Spatial Transformer: Learning Non-linear Spatial Interactions between Building Distribution and Pedestrian Flow

Shun Nakayama [1], Takahiro Kanamori [2,3], Wanglin Yan [4]

[1] School of Network and Information, Senshu University, Kawasaki, Japan - shunnkym@isc.senshu-u.ac.jp
[2] Research and Development Cernter, PASCO Corporation, Tokyo, Japan
[3] Keio Research Institute at SFC, Keio University, Fujisawa, Japan
[5] Faculty of Environment and Information Studies, Keio University, Fujisawa, Japan

**Abstract**

This study proposes a ring-based SpatialTransformer to learn how building uses at different distances from a railway station interact to generate pedestrian flow. Concentric ring buffers at 100-meter intervals up to 800 meters were defined around 100 randomly selected stations in Tokyo, treating each ring as a spatial token. Self-Attention was applied to learn inter-zone interactions directly from data, without prior structural assumptions. GPS-derived walking trip counts served as the target variable and Geographically Weighted Regression as the baseline. Across 30 independent trials, the SpatialTransformer consistently outperformed GWR in predictive accuracy. SHAP analysis revealed that mid-to-outer distance zone features dominate pedestrian flow prediction, while features from the 0-100m zone contributed little. The attention matrix showed that each distance zone attends most strongly to spatially distant zones, demonstrating that pedestrian flow is regulated by structural interactions across the entire catchment area rather than by any single zone in isolation. These findings challenge the compact city assumption that station-proximate development maximizes pedestrian flow, and suggest that land use distribution across the full walkable catchment area deserves greater consideration in urban planning practice.

## 1. Introduction

Japan is undergoing a period of population decline that demands a fundamental transformation of its urban structure. The Tokyo metropolitan area is projected to reach peak population in 2025, with the elderly already accounting for 23.0% of residents. As the working-age population shrinks and the proportion of older residents rises, shifting from car-dependent suburban sprawl toward compact, transit-oriented urban forms has become a pressing policy imperative. Such a transition is essential for preserving the mobility of an aging population, reducing infrastructure maintenance costs, and lowering environmental burdens. Against this backdrop, the Tokyo Metropolitan Government released guidelines for restructuring the city toward a more concentrated spatial form. The guidelines aim to create functional urban hubs around major railway stations by relaxing floor area restrictions, with the explicit goal of revitalizing pedestrian activity within walking distance. Underlying this policy is the assumption that the spatial distribution of building floor area and land use shapes how people move on foot. Yet the scientific basis for understanding which spatial configurations actually generate pedestrian flow remains insufficient.

In Japan, evidence that distance from a station does not predict commercial traffic volume in any simple monotonic way dates back at least to the 1990s (Yajima and Nakano, 1997). The assumption that pedestrian flow decreases uniformly with distance was shown early on to be an oversimplification. More recent studies have confirmed this complexity. An examination of large-scale development clusters in central Tokyo revealed that districts with similar building uses and floor areas nonetheless exhibited markedly different pedestrian movement patterns (Amaya et al., 2024). A GPS-based analysis at the block level further showed that trip generation rates vary substantially by building use and that these differences are modulated by locational conditions (Takeuchi et al., year). Taken together, these findings suggest that the relationship between building use and pedestrian flow cannot be explained by simple linear causality.

This concern is shared internationally. In recent years, machine learning has rapidly replaced ordinary least squares regression as the primary tool for uncovering nonlinear effects in the built environment. Using random forests, the relationship between building coverage ratio and pedestrian volume has been shown to be nonlinear: pedestrian volume increases with building density up to a threshold, beyond which it declines, and this effect is substantially modified by factors such as public transit accessibility and green space coverage (Zeng et al., 2024). A gradient boosted decision tree analysis of metro ridership in Shanghai revealed that commercial and leisure land uses exert the strongest influence on ridership, though the magnitude and direction of these effects vary across urban locations (Liu et al., 2023). In both cases, machine learning models substantially outperformed OLS and GWR in predictive accuracy, confirming that the relationship between the built environment and pedestrian flow is fundamentally nonlinear.

These approaches, however, share a common structural limitation. Random forests and gradient boosting treat each variable as an independent feature and estimate statistical interactions among them. Pedestrian flow in a station catchment area, however, is not determined by the building stock at any single location alone. What matters is the full spatial configuration: what uses are present close to the station, what uses exist at intermediate distances, and what lies at the outer edge of the walkable zone. Both model families can estimate the values of individual variables and their statistical interactions, but neither has any mechanism for learning how the use at one location relates to the use at another as a function of their spatial positions. These are fundamentally different questions.

OLS and GWR share this limitation. GWR accounts for spatial nonstationarity by estimating location-specific coefficients (Brunsdon et al., 1998), but it still represents relationships between variables as local linear approximations. The kind of interaction in which commercial uses near a station and residential uses farther away jointly generate pedestrian flow through their spatial relationship cannot be captured within this framework.

To address these limitations, this study proposes the application of the Transformer architecture (Vaswani et al., 2017) to spatial analysis. The Self-Attention mechanism learns, directly from data, how strongly each element in a sequence relates to every other element. Critically, it does not evaluate each element in isolation. It estimates the relational structure among elements. By representing concentric spatial zones around a station—each characterized by building floor area broken down by use category—as a sequence of spatial tokens, Self-Attention can learn how commercial uses at one distance zone interact with residential uses at another to generate pedestrian flow. This is fundamentally different from the statistical interaction terms estimated by conventional machine learning models.

This study addresses two research questions. First, what spatial patterns of building use distribution regulate pedestrian flow in station catchment areas? (RQ1) Second, what do these patterns reveal about the core assumption of compact city policy—that high-density development immediately adjacent to stations maximizes pedestrian flow? (RQ2) By answering these questions, this study aims to provide data-driven evidence to inform principles of density and land use distribution in urban planning.

## 2. Methodology

### 2.1 Study Area and Data

One hundred railway stations were randomly sampled from the 606 stations in the Tokyo metropolitan area. Random sampling was adopted to avoid geographic or route-specific bias and to ensure that the full diversity of station environments across the city was represented. The ratio of sample size to feature dimensionality is discussed in Section 2.2.

The dependent variable is the number of pedestrian trips within an 800-meter radius of each station, derived from individual GPS trajectory data collected in November 2021. Each trip is defined as an origin-destination pair, and only trips completed entirely within the 800-meter catchment area of each station are included. The 800-meter threshold follows the standard definition of walkable distance widely adopted in urban planning research and is appropriate for evaluating pedestrian activity within walking range. Trip counts were chosen as the measure of pedestrian flow because they capture purposeful movement rather than mere presence, and thus more directly reflect the spatial attractiveness of a station catchment area. Trip counts ranged from 111 to 29,847 per station, exhibiting a heavily right-skewed distribution. Left unaddressed, this skew would cause the model to optimize disproportionately toward low-volume stations. A log transformation (log1p) was therefore applied to both the features and the target variable, and a weighted loss function was introduced as described in Section 2.4.

Building data were drawn from the National Land Numerical Information building dataset. Each building polygon carries attributes including use category, footprint area, and total floor area. Building uses were consolidated into 15 categories. Total floor area was adopted as the primary metric because it captures building capacity rather than mere presence, allowing large office buildings and commercial complexes to be appropriately weighted relative to their potential to generate pedestrian flow.

### 2.2 Ring Buffer Zoning and Feature Construction

The feature design in this study rests on a single organizing premise: building uses at different distances from a station play different roles in generating pedestrian flow. To express this premise quantitatively, concentric spatial zones centered on each station were defined, and the total floor area by building use within each zone was used as the feature representation. A concentric, distance-based zoning scheme was adopted because the range of pedestrian movement around a railway station—a point facility—is governed more by distance from that point than by any particular direction.

Eight concentric ring buffers were generated around each station at 100-meter intervals up to 800 meters. The 800-meter limit follows the standard definition of a walkable catchment area widely adopted in urban planning research, and areas beyond this threshold are considered to have limited influence on the generation of walking trips. This procedure produced 800 polygons in total across 100 stations and 8 distance zones.

Assigning building floor area to zones by locating each building's representative point—such as its centroid—within a single zone introduces spatial misattribution for buildings that straddle zone boundaries. This problem is particularly acute for large buildings: a major mixed-use complex immediately adjacent to a station could be incorrectly assigned to a neighboring zone, undermining the spatial meaning of the features. Area-weighted apportionment was therefore applied. The intersecting area between each building polygon and each ring buffer polygon was computed, and the ratio of this intersection to the total footprint area of the building was multiplied by the total floor area to obtain the floor area attributed to each zone.

$$\text{allocated_tfa}_{i,r} = \text{TFA}_i \times \frac{A(B_i \cap R_r)}{A(B_i)} \quad (1)$$

Here $B_i$ denotes the polygon of building $i$, $R_r$ denotes the polygon of zone $r$, and $A(\cdot)$ denotes area. This procedure yielded 635,507 intersection records in total. The apportioned floor areas were then aggregated by station, distance zone, and building use, and rearranged into a feature matrix through a pivot transformation. The resulting feature dimensionality is 120 per station (8 zones × 15 building use categories). Of the 12,000 total elements, 3,401 (28.34%) are zero, but data coverage across all eight distance zones is 100% for every station, confirming that sparsity does not present a practical obstacle to analysis. The ratio of observations to feature dimensions (100 stations to 120 features, giving a ratio of 0.83) places this study within the small-data regime, a limitation addressed through multiple regularization mechanisms incorporated into the model.

### 2.3 SpatialTransformer Architecture

As argued in Section 1, what is needed is a mechanism that directly learns the interactions among building uses across multiple spatial zones as a function of their positions relative to the station. LSTMs process relationships between elements sequentially, making it difficult to capture long-range dependencies between arbitrary distance zones in parallel. Graph neural networks require the connection structure among nodes to be defined in advance, which prevents the model from freely learning which distance zones are related to which from the data alone. The Transformer's Self-Attention mechanism, by contrast, learns relationships among all tokens directly from data without any prior structural assumptions, making it the most appropriate choice for the questions this study addresses.

The model takes as input a three-dimensional tensor of shape $(N, 8, 15)$, where $N$ is the batch size, 8 is the number of distance zones, and 15 is the number of building use categories. Each 15-dimensional vector representing a distance zone is first projected into a 64-dimensional latent space through a linear transformation (feature embedding). A low-dimensional representation of 15 features is insufficient to capture latent relationships among different building uses. Projecting into a higher-dimensional space allows the model to represent not only the quantity of individual uses but also the latent patterns arising from their combinations. An embedding dimension of 64 was selected to balance representational capacity against parameter count, yielding a model scale that permits stable learning in a small-data setting.

Positional encoding is then added to make the physical location of each zone explicit. The Transformer architecture carries no inherent notion of token order, so without modification the zone at 100 meters and the zone at 700 meters would be indistinguishable, and all spatial meaning conveyed by distance would be lost. To resolve this, a sinusoidal positional encoding is applied, using the normalized center distance of each zone—expressed as a fraction of the maximum distance of 800 meters—as its basis.

$$\mathrm{PE}(r, 2j) = \sin!\left(\widetilde{d_r} \cdot 10000^{2j/d_{\text{model}}}\right) \tag{2}$$

$$\mathrm{PE}(r, 2j+1) = \cos!\left(\widetilde{d_r} \cdot 10000^{2j/d_{\text{model}}}\right) \tag{3}$$

Whereas the standard Transformer uses token index as the basis for positional encoding, this study uses physical distance. The model therefore begins training with geometric prior knowledge that adjacent zones are separated by equal intervals in space.

The sum of the embedding and positional encoding is passed through three layers of multi-head Self-Attention with eight heads. Each distance zone token queries all other tokens to determine which zones carry the most relevant information for estimating its contribution to pedestrian flow, then aggregates information according to the resulting weights. Running eight such queries in parallel allows the model to simultaneously learn different types of spatial relationship—complementary interactions between adjacent zones, synergistic effects between the station-proximate zone and the outer edge, and so on. The choice of eight heads follows from dividing the embedding dimension (64) by the head count, yielding eight dimensions per head, which provides the minimum representational capacity required for each head. Three layers were used so that the first layer captures local relationships between neighboring zones, while the second and third layers abstract progressively broader spatial patterns. Each attention layer is followed by a residual connection, layer normalization, a feed-forward network (hidden dimension 128, GELU activation, dropout $p = 0.2$), a second residual connection, and a second layer normalization. GELU was preferred over ReLU because it is less susceptible to gradient vanishing and is known to support stable training in Transformer architectures.

The eight tokens produced by three attention layers are aggregated into a single 64-dimensional vector through global average pooling. Average pooling was chosen over max pooling to integrate information from all eight distance zones equally, without privileging any single zone. This is consistent with the premise that pedestrian flow arises from the cooperative structure of building use across the entire catchment area, rather than from any single outstanding feature. The aggregated vector is passed through a nonlinear transformation layer and then through a four-layer regression head (each layer incorporating layer normalization, GELU activation, and dropout) to produce a one-dimensional prediction of trip count. All weights were initialized using Xavier uniform initialization, which scales the variance of activations according to the input and output dimensions of each layer, thereby suppressing vanishing and exploding gradients in deep networks. The total parameter count of the model is approximately 49,000.

## 2.4 Model Training and Evaluation

Huber loss with $\delta = 1.0$was adopted as the loss function. Mean squared error is overly sensitive to outliers and is therefore poorly suited to right-skewed distributions such as the trip count data used here. Mean absolute error is robust to outliers but produces a constant gradient in regions of small error, which can destabilize convergence. Huber loss combines the properties of both: it behaves as squared error when the residual is below $\delta$and as absolute error when the residual exceeds $\delta$. An additional sample-weighting scheme was applied to assign greater loss to high-flow stations: those above the 75th percentile of the training distribution received a weight of 1.5, and those above the 90th percentile received a weight of 2.0. Without this weighting, the model risks over-optimizing toward the large number of low-flow stations at the expense of the small number of high-flow stations that dominate the right tail of the distribution. The weight values were set empirically to increase attention to high-flow stations without inducing overfitting.

Adam optimization was used with a learning rate of 0.001 and weight decay $\lambda = 10^{-4}$. Adam was selected because its adaptive adjustment of first and second gradient moments reliably produces stable convergence without manual tuning of the learning rate, a property well established across deep learning research. Weight decay functions as L2 regularization and contributes to suppressing overfitting in the small-data setting. A ReduceLROnPlateau scheduler was applied to reduce the learning rate by a factor of 0.7 whenever validation loss failed to improve for 15 consecutive epochs, encouraging convergence toward better local optima when progress stalls. Early stopping was triggered if validation loss showed no improvement for 50 consecutive epochs, with a maximum of 500 epochs. Early stopping serves both to prevent overfitting and to avoid unnecessary computation.

The dataset was split into 80 training stations and 20 test stations through random partitioning. A different random seed was used for each of the 30 trials, so that each trial operates on an independent random split. This design captures not only the variability due to random weight initialization but also the variability introduced by different data partitions. The trial yielding the highest test $R^2$was selected as the final model, and its partition was carried forward into the subsequent attention weight analysis and SHAP analysis. All evaluation metrics were computed in the log1p-transformed and standardized space. Evaluating in the original trip count scale was avoided because prediction error is systematically larger for high-flow stations, causing RMSE and MAE to be dominated by a small number of outlying observations. Reporting metrics in the transformed space ensures that predictive accuracy is assessed consistently across stations regardless of their absolute flow magnitude.

GWR was adopted as the comparison baseline. It accounts for spatially varying relationships by estimating location-specific regression coefficients (Brunsdon et al., 1998) and represents the standard approach against which this study's method is benchmarked. The full 120-dimensional feature set cannot be passed directly to GWR: when the number of dimensions (120) exceeds the number of observations (100 stations), the local regression matrices become singular or near-singular, making stable estimation impossible. A 15-dimensional variable set was therefore constructed by aggregating floor area across building uses within each distance zone and used as input to GWR. This asymmetry in input dimensionality—15 dimensions for GWR versus 120 for the SpatialTransformer—is not a methodological flaw but rather reflects a contribution of this study: the SpatialTransformer's regularization mechanisms enable stable learning under high-dimensional, small-sample conditions in which GWR cannot function, and allow the model to exploit detailed spatial information that GWR must discard. To ensure a fair comparison, both models were evaluated on the same log1p-transformed and standardized scale.

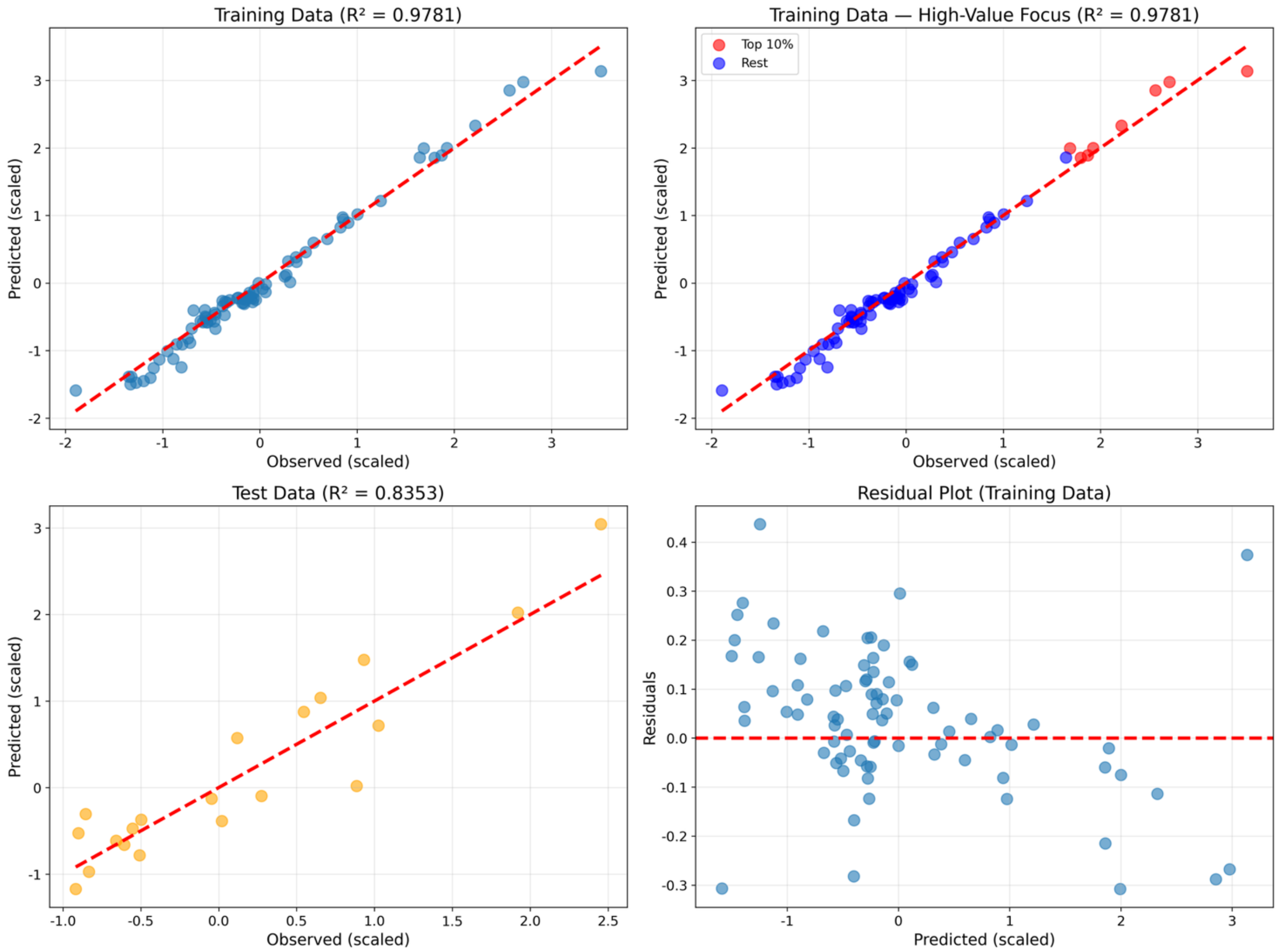

Figure 1 Scatter plots of predicted versus observed values and residual plots for the final model (training and test data).

Two complementary analyses were conducted to interpret the model. The first averaged the multi-head attention weights of the final layer across all heads to produce an importance score for each distance zone. Attention weights indicate how information is aggregated rather than how much each feature contributes to the prediction, and their interpretability as measures of feature importance remains contested in the literature (Jain and Wallace, 2019). This analysis is therefore treated as a supplementary interpretive tool. The second analysis used SHAP (Lundberg and Lee, 2017) to quantify the contribution of each of the 120 individual features to the pedestrian flow prediction. Grounded in Shapley values from cooperative game theory, SHAP fairly distributes the marginal contribution of each feature to the predicted outcome and compensates for the interpretive limitations of attention weights. Fifty samples drawn randomly from the training data served as the background dataset, and SHAP values were computed for all 20 test samples.

## 3. Results

### 3.1 Predictive Accuracy

Predictive accuracy was compared between the SpatialTransformer and GWR in the standardized space (after log1p transformation and StandardScaler normalization). Both models were evaluated across 30 independent trials, each using a different random train/test split, with predictive accuracy assessed on 20 test stations per trial.

Across 30 trials, GWR achieved a mean test $R^2$ of $-5.486\pm5.209$, RMSE of $2.259\pm0.517$, and MAE of $1.668\pm0.357$. The best trial (Trial 15) yielded a test $R^2$ of $-0.382$, RMSE of 1.654, and MAE of 1.346. Across the same 30 trials, the SpatialTransformer achieved a mean test $R^2$ of $0.152\pm0.652$, RMSE of $0.822\pm0.283$, and MAE of $0.521\pm0.144$. The best trial (Trial 20) yielded a test $R^2$ of 0.835, RMSE of 0.381, and MAE of 0.317, and was selected as the final model. On the training data, the final model achieved $R^2$=0.978, RMSE=0.148, and MAE=0.113.

The SpatialTransformer outperformed GWR in both the best trial and the 30-trial average. In the best trial, the difference in test $R^2$ between the two models was 1.217 points (0.835 versus $-0.382$). However, the 30-trial average test $R^2$ of the SpatialTransformer ($0.152\pm0.652$) also shows substantial variability across trials, indicating that stable generalization was not consistently achieved. It should also be noted that the trial with the highest evaluation score during training does not necessarily coincide with the trial yielding the best test performance across the 30 trials.

Figure 1 presents scatter plots of predicted versus observed values alongside residual plots, arranged in four panels. In the scatter plots, the horizontal axis shows observed values and the vertical axis shows predicted values; points closer to the diagonal reference line indicate more accurate predictions. In the residual plots, the horizontal axis shows predicted values and the vertical axis shows residuals. On the training data, predicted values follow observed values closely. On the test data, prediction error tends to increase, particularly for high-flow stations.

### 3.2 Attention Weight Analysis

Importance scores for the eight distance zones were computed by averaging the attention weights of the final layer across all heads and then across all 100 stations. Figure 2 shows the relationship between distance from the station and attention weight as a line graph. The horizontal axis represents distance from the station in meters and the vertical axis represents attention weight. The score peaks at the 0-100m zone, drops sharply through the 100-200m zone, reaches its minimum at 300-400m, recovers slightly at 400-500m, and remains approximately flat from 500m to 800m.

The 8×8 attention matrix shows which zone each distance zone attends to most strongly. The 0-100m zone attends most strongly to the 700-800m zone (weight=0.191); the 100-200m zone to the 500-600m zone (0.200); the 200-300m zone to the 400-500m zone (0.176); the 300-400m zone to the 700-800m zone (0.173); the 400-500m zone to the 700-800m zone (0.147); the 500-600m zone to the 700-800m zone (0.176); the 600-700m zone to the 0-100m zone (0.171); and the 700-800m zone to the 400-500m zone (0.177). Diagonal entries, representing each zone's attention to itself, ranged from 0.074 to 0.165, with off-diagonal attention predominating throughout. It should be noted that attention weights depend on model parameters and may vary across trials. This analysis is therefore treated as a supplementary interpretive tool that complements the SHAP analysis presented in the following section.

### 3.3 SHAP Analysis

SHAP values were computed for all 20 test stations across the full set of 120 features (8 distance zones × 15 building use categories). The mean absolute SHAP value (Mean |SHAP|) for each feature indicates the magnitude of its contribution to the pedestrian flow prediction, with larger values reflecting stronger influence.

Figure 3 presents a bar chart of SHAP feature importance. The horizontal axis shows Mean |SHAP| and the vertical axis lists feature names expressed as combinations of distance zone and building use category. Features are ranked in descending order of importance, with longer bars indicating greater contribution to the prediction.

The ten most important features, in descending order, were: 700-800m_Commercial (0.0810), 400-500m_Commercial (0.0619), 700-800m_Industrial Factory (0.0515), 400-500m_Government (0.0495), 100-200m_Education/Culture (0.0484), 200-300m_Industrial Factory (0.0469), 100-200m_Commercial (0.0446), 700-800m_Lodging/Entertainment (0.0438), 200-300m_Government (0.0422), and 700-800m_Healthcare/Welfare (0.0412). Among the top 20 features, the 700-800m zone accounted for five entries, the 400-500m zone for four, and the 200-300m and 100-200m zones for three each. The 0-100m zone appeared only once, represented by detached housing.

Figure 4 presents a beeswarm summary plot. The horizontal axis shows the SHAP value for each observation and the vertical axis lists feature names. Each point corresponds to one station. Point color encodes the magnitude of the feature value: red indicates high values (large floor area) and blue indicates low values (small floor area). Points to the right of zero indicate a positive contribution to the predicted trip count, while points to the left indicate a negative contribution. For example, where red points for 700-800m_Commercial cluster to the right, stations with larger commercial floor area in that zone are predicted to generate more trips.

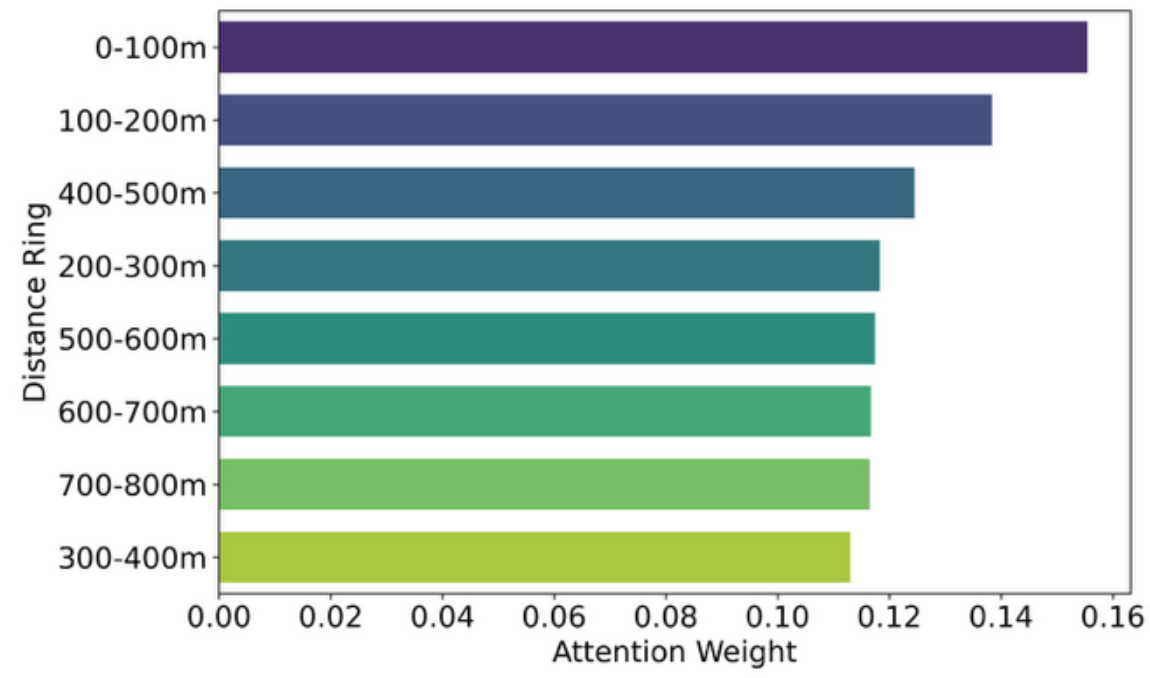

Figure 2 Mean attention weight by distance from station.

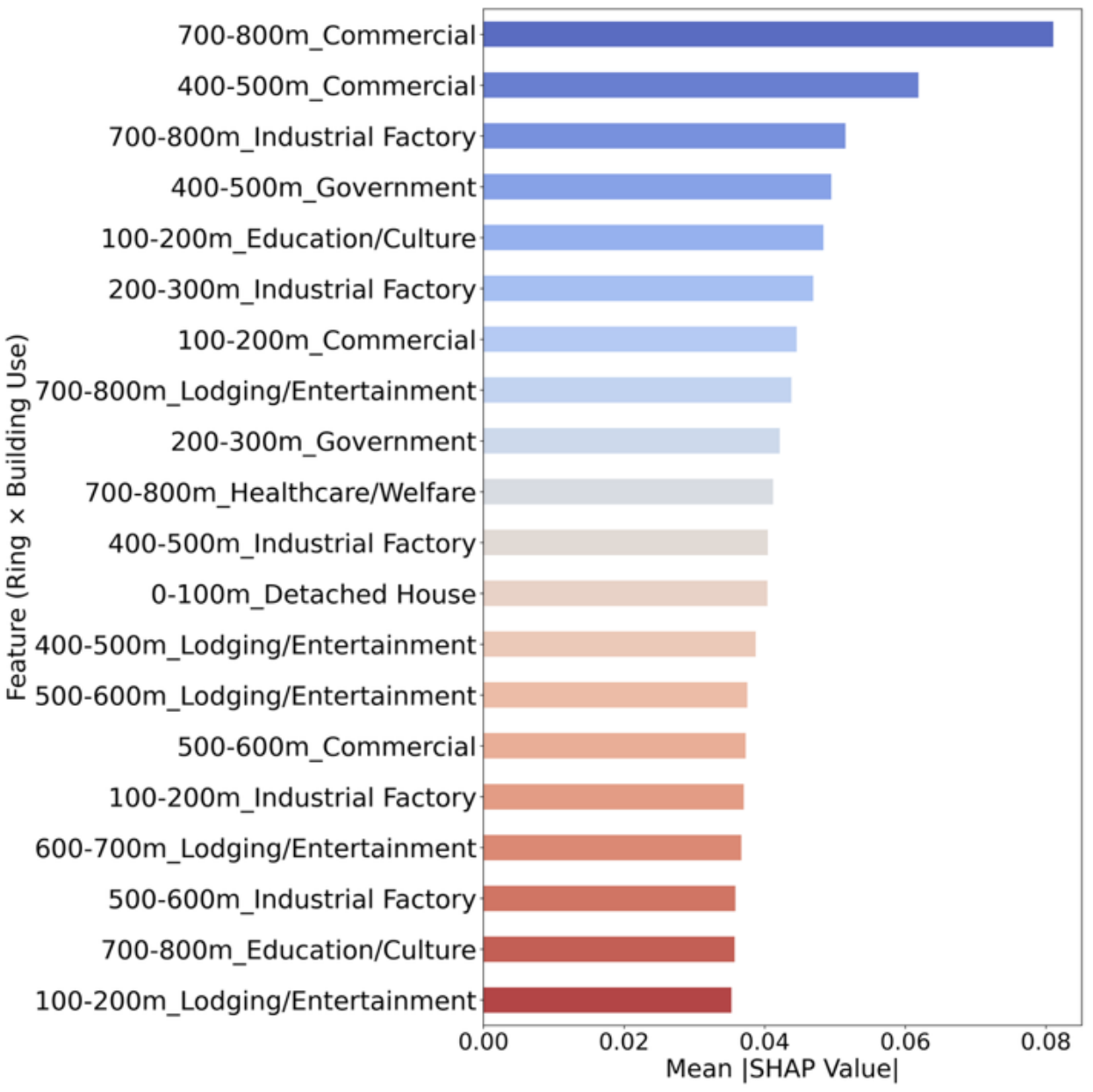

Figure 3 Top 20 features ranked by mean absolute SHAP value.

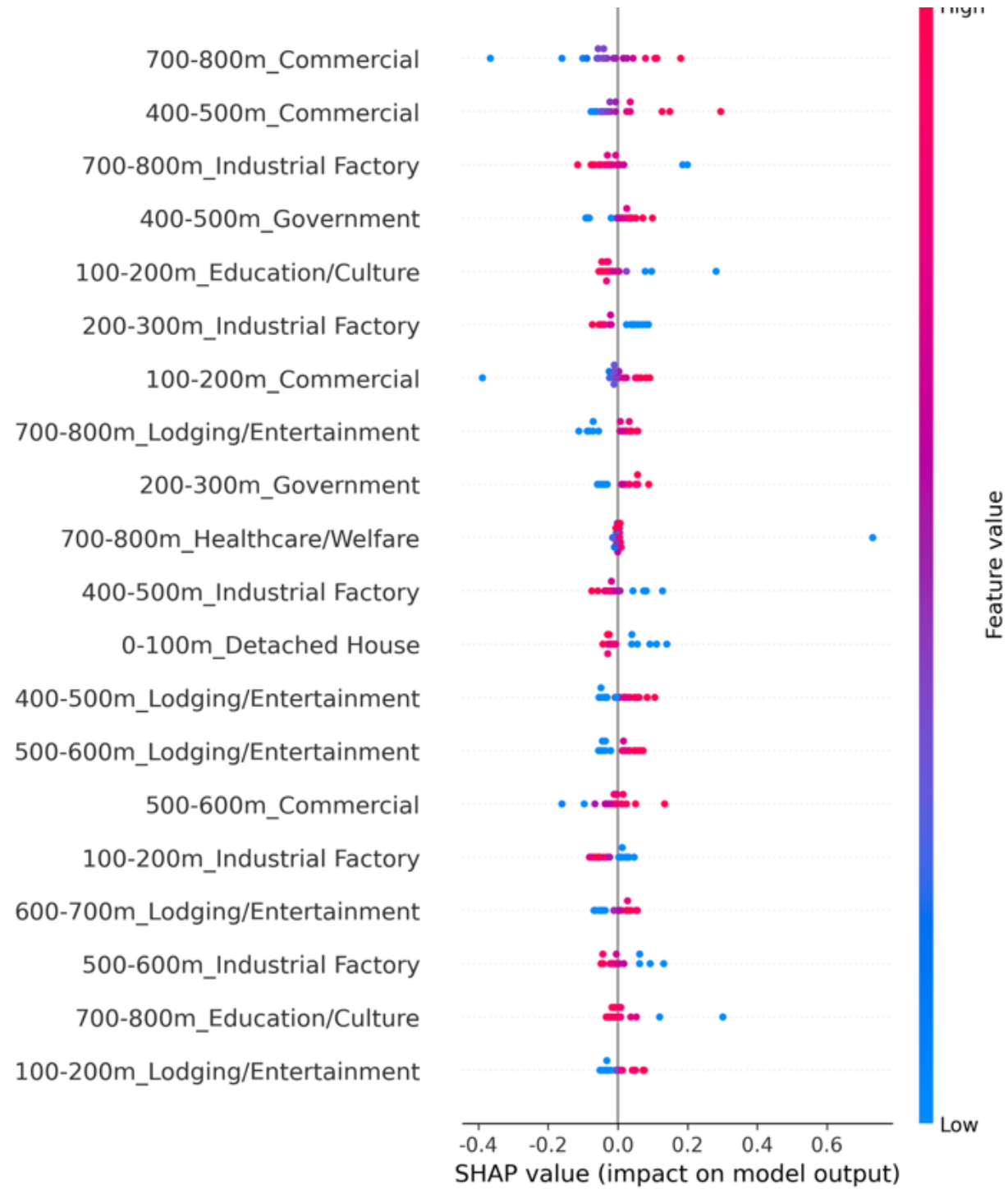

Figure 4 SHAP summary plot for the top 20 features.

## 4. Discussions

### 4.1 Predictive Performance of the SpatialTransformer

In the best trial, the SpatialTransformer achieved a test $R^2$ of 0.835, substantially outperforming GWR, whose best trial yielded a test $R^2$ of −0.382 under the same evaluation conditions. Across all 30 trials, the SpatialTransformer's mean test $R^2$ (0.152±0.652) also exceeded that of GWR (−5.486±5.209), indicating a consistent advantage across both metrics.

However, the 30-trial mean test $R^2$ of 0.152±0.652 shows considerable variability, and stable generalization was not consistently achieved. This instability arises from three compounding factors. The first is the small test set size. With only 20 test stations per trial, the composition of the test set varies substantially across trials. The total variance of trip counts in the test set—the denominator of $R^2$—fluctuates dramatically depending on which 20 stations are selected. Trip counts across the 100 stations range from 111 to 29,847, a factor of approximately 270, meaning that the inclusion or exclusion of a single high-flow station can change the denominator by an order of magnitude. This is a statistical sampling problem that is independent of model quality. The second factor is the ratio of feature dimensions to sample size. With 120 features and only 80 training stations, the risk of overfitting remains high despite the regularization mechanisms built into the model. The gap between the mean training $R^2$ (0.978) and the mean test $R^2$ (0.152) reflects this directly. The third factor is the mismatch between the model selection criterion and test performance. The trial selected as the final model based on its training-phase evaluation score does not necessarily coincide with the trial yielding the best test performance across the 30 trials.

These problems can be substantially mitigated by expanding the analysis to all 606 railway stations in Tokyo. A full-dataset analysis would yield a test set of n=121, largely resolving the first problem. The ratio of training samples to feature dimensions would rise to 4.0, greatly reducing the risk of overfitting. The third problem, however, is independent of sample size and requires a separate reconsideration of the model selection criterion. These remain as priorities for future work.

The asymmetry in input dimensionality between the two models also warrants attention. GWR was given a 15-dimensional aggregated feature set due to the instability of local regression under high-dimensional, small-sample conditions, whereas the SpatialTransformer was given the full 120-dimensional feature set. The difference in predictive accuracy therefore reflects not only architectural differences but also a difference in the amount of spatial information available to each model. Seen from another angle, the ability of the SpatialTransformer to exploit detailed spatial information that GWR cannot handle—through its built-in regularization mechanisms—represents a methodological contribution of this study.

### 4.2 Spatial Patterns of People Flow and Policy Implications

RQ1 asked what spatial patterns of building use distribution regulate pedestrian flow in station catchment areas. RQ2 asked what those patterns reveal about the assumption underlying compact city policy—that high-density development immediately adjacent to stations maximizes pedestrian flow.

The non-diagonal attention structure revealed by the attention matrix provides the most direct answer to RQ1. For example, the 0-100m zone attends most strongly to the 700-800m zone, meaning that the nearest token draws on information from the outermost zone when estimating its contribution to pedestrian flow. GWR represents relationships between variables as local linear approximations at each observation point and therefore cannot model a structure in which building uses at one distance zone interact with building uses at a spatially distant zone. The SpatialTransformer learns this structure directly from data through Self-Attention. Pedestrian flow in station catchment areas is therefore regulated not by the building stock of any single zone in isolation, but by a structural pattern in which multiple distance zones mutually reference one another. This is a finding that GWR and conventional machine learning approaches cannot capture, and it constitutes the methodological core of this study's contribution.

The attention weight analysis elaborates on this pattern. The importance score for the 0-100m zone (0.155) is the highest among all eight zones, but the score drops to its minimum at 300-400m (0.113) before rising again at 400-500m (0.124), the third-highest value. This non-monotonic pattern suggests that both the station-proximate zone and the mid-distance zone (400-500m) carry relatively high importance, a structure that cannot be characterized as simple distance decay.

The SHAP analysis gives this pattern a more concrete form. The most important feature is 700-800m Commercial (0.0810), followed by 400-500m Commercial (0.0619). Only one feature from the 0-100m zone—detached housing—appeared among the top 20. Mid-to-outer distance zones dominate the prediction. A notable finding is that industrial factories and government facilities rank highly across multiple distance zones: 700-800m Industrial Factory (3rd, 0.0515), 400-500m Government (4th, 0.0495), 200-300m Industrial Factory (6th, 0.0469), and 200-300m Government (9th, 0.0422) all appear in the top ten. These land uses generate large volumes of trips through mechanisms distinct from commercial facilities. Industrial factories produce sustained trip volumes through employee commuting as well as regular visits by delivery and logistics workers. Government facilities generate a steady stream of trips from residents visiting for administrative purposes. Both use types share the characteristic of producing concentrated, repetitive trips at specific times of day, suggesting that they serve as a stable foundation for pedestrian flow in station catchment areas. This finding complements perspectives that attribute pedestrian flow primarily to commercial agglomeration.

These results provide a clear answer to RQ2. The assumption that building uses immediately adjacent to the station most strongly regulate pedestrian flow is not supported. This finding is consistent with the early observation that no clear monotonic relationship exists between station distance and traffic volume for commercial facilities (Yajima and Nakano, 1997), and aligns with the finding that districts with similar building uses and floor areas nonetheless exhibit markedly different pedestrian movement patterns (Amaya et al., 2024). The policy implication is that designing the distribution of land uses across the entire walkable catchment area may be more consequential than concentrating development immediately adjacent to the station. It should be noted, however, that this study demonstrates the relative magnitude of contributions and does not imply that station-proximate development is unnecessary. Given the substantial variability in predictive accuracy across trials, these findings should be regarded as exploratory patterns requiring verification through analysis of a larger station sample.

### 4.3 Discrepancy Between Attention Weights and SHAP

A discrepancy exists between the results of the attention weight analysis and the SHAP analysis. The attention weight analysis identifies the 0-100m zone as the most important, whereas the SHAP analysis shows that features from the 0-100m zone contribute little to the predicted values. The zone most strongly referenced by the model does not appear as a meaningful contributor to prediction.

The cause of this discrepancy is not clear at this stage. One possible explanation is that attention weights indicate the degree to which information from each zone is referenced, rather than the contribution of each feature to the predicted outcome (Jain and Wallace, 2019). Attention and SHAP may therefore be measuring different aspects of the model, and it is in principle possible for them to yield different results. What remains unexplained, however, is why the zone most strongly attended to contributes so little to the final prediction. This study cannot offer a definitive answer to that question.

The issue is closely connected to unresolved questions in Transformer interpretability research. It has already been shown that attention weights do not directly reflect predictive contributions (Jain and Wallace, 2019), and debate continues over what metrics can appropriately characterize the internal representations of Transformer models. By applying both attention weight analysis and SHAP analysis in a spatial analytic context, this study demonstrates concretely that this interpretability challenge extends to GeoAI applications. For the purpose of quantifying feature contributions to prediction, SHAP analysis—grounded in the game-theoretic axioms of Shapley values—is the more reliable basis for interpretation. Developing a framework for interpreting the spatial meaning of Transformer attention structures remains an important direction for future research.

## 5. Conclusion

This study addressed two research questions concerning pedestrian flow in station catchment areas. RQ1 asked what spatial patterns of building use distribution regulate pedestrian flow. RQ2 asked what those patterns reveal about the assumption underlying compact city policy—that high-density development immediately adjacent to stations maximizes pedestrian flow. Existing approaches, including OLS, GWR, and machine learning models such as random forests and gradient boosting, can estimate statistical interactions among building environment variables but lack a mechanism for learning how building uses at different spatial positions relate to one another. To fill this gap, this study proposed a ring-based SpatialTransformer that treats concentric distance zones around each station as spatial tokens and applies Self-Attention to learn inter-zone interactions directly from data.

The model was applied to 100 randomly selected railway stations in Tokyo. Building floor area by use category was aggregated into eight concentric ring buffers at 100-meter intervals up to 800 meters, yielding a 120-dimensional feature set per station. GWR was adopted as the baseline. To ensure a fair comparison, both models were evaluated in the same standardized space after log1p transformation. Across 30 independent trials with random train/test splits, the SpatialTransformer consistently outperformed GWR in both the best trial ($R^2$=0.835 versus −0.382) and the 30-trial mean (0.152±0.652 versus −5.486±5.209). However, the large variability in test performance across trials reflects the instability of applying a high-dimensional model to a small sample of 100 stations, and the findings should be regarded as exploratory pending verification with a larger dataset.

With respect to RQ1, the attention matrix revealed a non-diagonal reference structure in which each distance zone attends most strongly to zones other than itself. This demonstrates that pedestrian flow is regulated not by the building stock of any single zone in isolation, but by a structural pattern in which multiple distance zones mutually reference one another—a finding that GWR and conventional machine learning approaches cannot produce. SHAP analysis further showed that mid-to-outer distance zone features dominate pedestrian flow prediction, with 700-800m Commercial ranking first and 400-500m Commercial ranking second. Industrial factories and government facilities appeared prominently across multiple distance zones, reflecting their role in generating concentrated, repetitive trips through commuting, logistics, and administrative visits—a finding that complements perspectives attributing pedestrian flow primarily to commercial agglomeration.

With respect to RQ2, features from the 0-100m zone were largely absent from the top SHAP rankings, and no monotonic relationship was found between proximity to the station and feature importance. These results do not support the assumption that station-proximate development most strongly regulates pedestrian flow. The policy implication is that designing land use distribution across the entire walkable catchment area warrants greater attention than concentrating development at the station front alone.

Finally, a discrepancy was identified between the attention weight analysis and the SHAP analysis: the zone most strongly referenced by the model contributed little to the predicted values as measured by SHAP. The cause remains unresolved and points to broader open questions in Transformer interpretability research. Developing a framework for interpreting the spatial meaning of attention structures in GeoAI applications is an important direction for future work, alongside expanding the analysis to the full set of 606 stations in Tokyo to obtain more stable and generalizable results.

## Acknowledgement

This study was conducted as part of the Tokyo Metropolitan Government's "Smart Tokyo Advanced Case Creation Project Led by Local Communities," specifically the collaborative project "Co-creation of Digital Area Design in Oimachi."